\pdfoutput=1

\documentclass[11pt]{article}

\usepackage{acl}

\usepackage{times}
\usepackage{latexsym}
\usepackage{graphicx}
\usepackage{import}
\usepackage{enumitem}
\usepackage{multirow}
\usepackage{subfig}
\usepackage{booktabs}
\usepackage{algpseudocode}
\usepackage{algorithm}

\usepackage{multicol}
\usepackage{amsmath}
\usepackage{amsfonts}
\usepackage{enumitem}
\usepackage{arydshln}
\usepackage{todonotes}
\usepackage{sidecap}
\usepackage{soul}
\usepackage{lipsum}
\newcommand\blfootnote[1]{%
  \begingroup
  \renewcommand\thefootnote{}\footnote{#1}%
  \addtocounter{footnote}{-1}%
  \endgroup
}

\usepackage[T1]{fontenc}

\usepackage[utf8]{inputenc}

\usepackage{microtype}

\usepackage{inconsolata}

\newcommand{\sbic}{\texttt{SBIC}}
\newcommand{\latent}{\texttt{LatentHatred}}
\newcommand{\concept}{\texttt{ConceptNet}}
\newcommand{\stereo}{\texttt{StereoKG}}
\newcommand{\toxbart}{\texttt{Tox-BART}}
\newcommand{\toxbartfull}{\texttt{toxicity-infused-BART}}

\title{\texttt{Tox-BART}: Leveraging Toxicity Attributes for Explanation Generation of Implicit Hate Speech}

\author{Neemesh Yadav$^{*\dagger}$, 
Sarah Masud$^{*\dagger}$,\\ 
  \textbf{Vikram Goyal}$^\dagger$, \textbf{Md Shad Akhtar}$^\dagger$,
  \textbf{Tanmoy Chakraborty}$^\ddagger$\\
  $^\dagger$IIIT Delhi, $^\ddagger$IIT Delhi\\
\texttt{\{neemesh20529,sarahm,vikram,shad.akhtar\}@iiitd.ac.in, tanchak@iitd.ac.in}}

\begin{document}
\maketitle
\begin{abstract}
Employing language models to generate explanations for an incoming implicit hate post is an active area of research. The explanation is intended to make explicit the underlying stereotype and aid content moderators. The training often combines top-k relevant knowledge graph (KG) tuples to provide world knowledge and improve performance on standard metrics. Interestingly, our study presents conflicting evidence for the role of the \emph{quality} of KG tuples in generating implicit explanations. Consequently, simpler models incorporating external toxicity\footnote{\color{red}{\noindent\textbf{Disclaimer:} The paper contains examples of hateful speech included solely for contextual understanding.}} signals outperform KG-infused models. Compared to the KG-based setup, we observe a comparable performance for \sbic\ (\latent) datasets with a performance variation of +$0.44$ (+$0.49$), +$1.83$ (-$1.56$), and -$4.59$ (+$0.77$) in BLEU, ROUGE-L, and BERTScore. Further human evaluation and error analysis reveal that our proposed setup produces more precise explanations than zero-shot GPT-3.5, highlighting the intricate nature of the\blfootnote{* Equal Contribution} task.
\end{abstract}

\section{Introduction}
\begin{figure}[!t]
\includegraphics[width=\columnwidth]{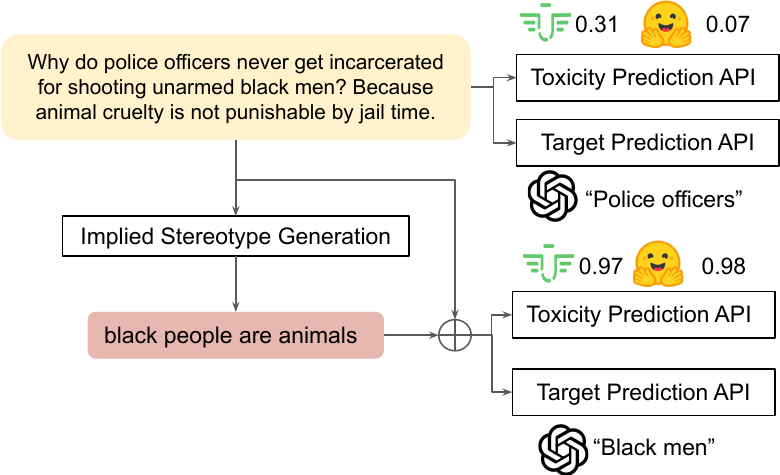}
\centering
  \caption{A sample text (verbatim from \sbic) witnessing an improvement in toxicity and target detection when the incoming post is infused with implied context. We infer toxicity scores from the Unitary toxicity API and Toxigen-RoBERTa. For target detection, we prompt the ChatGPT user interface.}
  \label{fig:motivation}
\vspace{-2mm}
\end{figure}

Despite subjectivity, hate speech can be characterized as ``\emph{communication that humiliates and denigrates a group or an individual based on their identity}" \cite{nockleby2000}. Implicit hate speech, in particular, uses circumlocution and stereotyping to mask the hate \cite{gao-etal-2017-recognizing}, which content moderation systems (human or computer-aided) sometimes fail to understand. We observe this even with sophisticated systems like ChatGPT (GPT-3.5). The system's efficacy improves when the implicit hate is accompanied by its underlying explicit explanation as outlined in Figure \ref{fig:motivation}. However, elucidating the underlying implied hate is a non-trivial task. It requires cognizance of societal norms \cite{forbes-etal-2020-social}, world knowledge \cite{lin-2022-leveraging}, contextual reasoning \cite{zhou-etal-2023-cobra}, etc. Recently proposed systems build upon the ability of Large Language Models (LLMs) to sufficiently capture and generate explanations for implicit content \cite{mun-etal-2023-beyond, zhang-etal-2023-biasx}. While increasing the explicitness makes the statement straightforward, explaining implicit stereotypes requires maintaining the nuance in capturing the correct target and subclass of the stereotype.

\textbf{Infusing Knowledge Signals.} Despite the rising trend in using LLMs, experimenting with them is prohibitive due to resource constraints. Hence, this work focuses on Pretrained Language models (PLMs), especially building upon MIXGEN \cite{sridhar-yang-2022-explaining} - a Knowledge Graph (KG) infused BART-based model \cite{lewis-etal-2020-bart}. PLMs are often augmented with KG tuples \cite{sridhar-yang-2022-explaining,chang-etal-2020-incorporating, lin-2022-leveraging} to enhance the model's reception of world knowledge, as reported by improved performance metrics. Yet, we hypothesize that 
\emph{the process of obtaining KG tuples is task agonistic and may not account for the multi-hop/indirect nature of hate.} \citet{lin-2022-leveraging} have made similar observations in their use of Wikipedia entities for classifying the type of implicit hate. It prompts us to empirically examine -- \textit{``the impact of the \emph{quality} of KG tuples on the performance of PLMs for implicit hate explanation."}

Working with two publicly available implicit hate datasets -- \sbic\ \cite{sap-etal-2020-social} and \latent\ \cite{elsherief-etal-2021-latent}, and KGs -- \concept\ \cite{conceptnet5-speer} and \stereo\ \cite{deshpande-etal-2022-stereokg}, we observe that (Table \ref{table:quality_cos_results}) \emph{replacing the top-k most relevant tuples with either the bottom-k least relevant tuples or random-k tuples does not necessarily cause the relative performance to languish}. In Section \ref{section:cos_quality_relevance}, we deep dive into this anomaly and perform a two-part error analysis inspecting the retrieval and manual scores. Succinctly speaking, our investigation corroborates our hypothesis.

\begin{figure}[!t]
\centering
\includegraphics[width=\columnwidth]{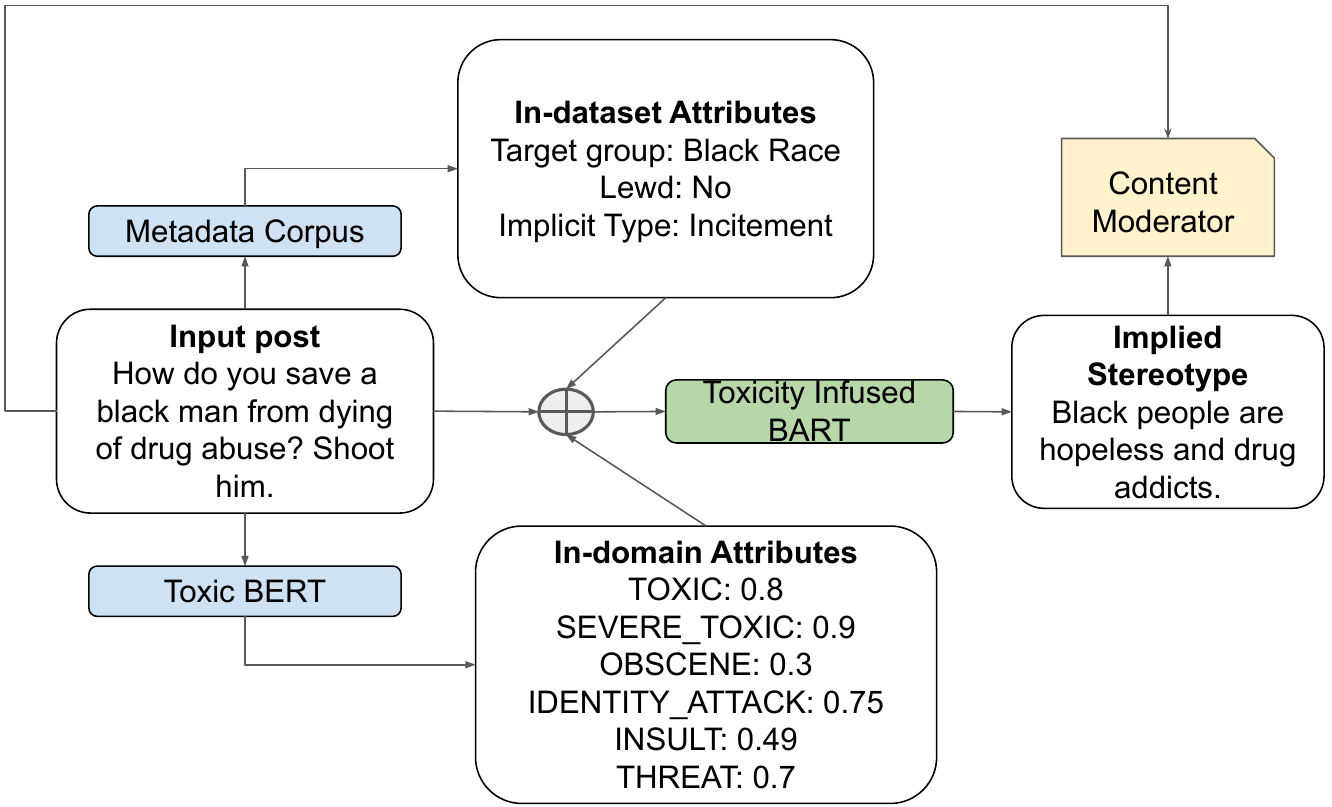}
  \caption{Workflow of our proposed system \toxbart\ utilizing toxicity attributes (\textit{in-dataset} and \textit{in-domain}) for explaining implicit hate.}
  \label{fig:sample_fig}
\end{figure}

\textbf{Proposed Methodology.} Looking beyond KG-infusion, we seek \textit{``what alternate signals can be leveraged to enrich the explanation of implied stereotypes?''} To this end, we investigate the infusion of ``toxicity attributes.'' These ``toxicity attributes'' \cite{alkhamissi-etal-2022-token} can be defined as indicators outside the post text that convey the power dynamics \cite{zhou-etal-2023-cobra}, target groups \cite{sap-etal-2020-social}, insult-type \cite{elsherief-etal-2021-latent} or hate intensity \cite{10.1145/3534678.3539161} regarding the post. We broadly classify them as -- \textit{in-dataset} or \textit{in-domain}. The former is obtained from the auxiliary annotations (about the speaker, target, etc.) already available in the given dataset. Meanwhile, our \textit{in-domain} signals enlist toxicity indicators obtained by finetuning a BERT regressor on the Jigsaw dataset \cite{jigsaw-unintended-bias-in-toxicity-classification}. 

As outlined in Figure \ref{fig:sample_fig}, we then employ ``toxicity attributes'' to formulate a BART-based model \textit{aka} \toxbart\ to generate implied explanations. A ``metadata corpus'' is the post's (incoming data points) complementary information. For example, if the post comes from Twitter, then the likes and reply count \cite{10.1145/3292522.3326028} becomes engagement-based metadata features. Other times, this information can be completely unsupervised/unlabelled but still functional, like the user's ego network \cite{kulkarni2023revisiting}.

Compared to the KG-based setup, we observe a comparable performance for \sbic\ (\latent) datasets with a performance variation of +$0.44$ (+$0.49$), +$1.83$ (-$1.56$), and -$4.59$ (+$0.77$) in BLEU, ROUGE-L, and BERTScore. We also look into how varying the quality of toxicity signals leads to the expected loss in performance, which is another indicator of the consistency of \toxbart. In addition, we inspect the role of zero-shot prompted GPT-3.5\footnote{gpt-3.5-turbo}. Based on standard metrics, human evaluation, and error analysis, we observe that \toxbart\ outperforms GPT-3.5 by producing more specific explanations. 

\textbf{Contributions.} To summarise, this work\footnote{Code: \url{https://github.com/LCS2-IIITD/TOXBART}}:
\begin{itemize}[noitemsep, nolistsep, topsep=0pt, leftmargin=1em]
    \item Proposes the infusion of ``toxicity attributes'' via \toxbart\ (Section \ref{sec:tox_bart}). The study showcases that infusion of toxicity signals is at par with KG-infusion for explanation generation (Section \ref{sec:exp_rq2}).
    \item Assess that \toxbart\ can generate more specific explanations than GPT-3.5 by performing human evaluation and error analysis (Section \ref{sec:exp_rq2}).
    \item Empirically establishes that ``richness/relevance'' of the KG tuples has little to no difference for implied explanation generation (Section \ref{section:cos_quality_relevance}). These findings have far-reaching implications for adopting KG in subjective tasks.
\end{itemize}
\emph{Through extensive ablations on the toxicity and the KG attributes, we register a higher sensitivity in performance by toxicity signals. Compared to KG signals, toxicity signals are more sensitive to subtle changes in the input post, making them superior contextual information when working with subtle setups like implicit hate speech. Further, the results of ``in-dataset'' attributes reinstate the importance of human labeling in the hate moderation pipeline.}  

\begin{SCfigure*}
\includegraphics[width=0.65\textwidth]{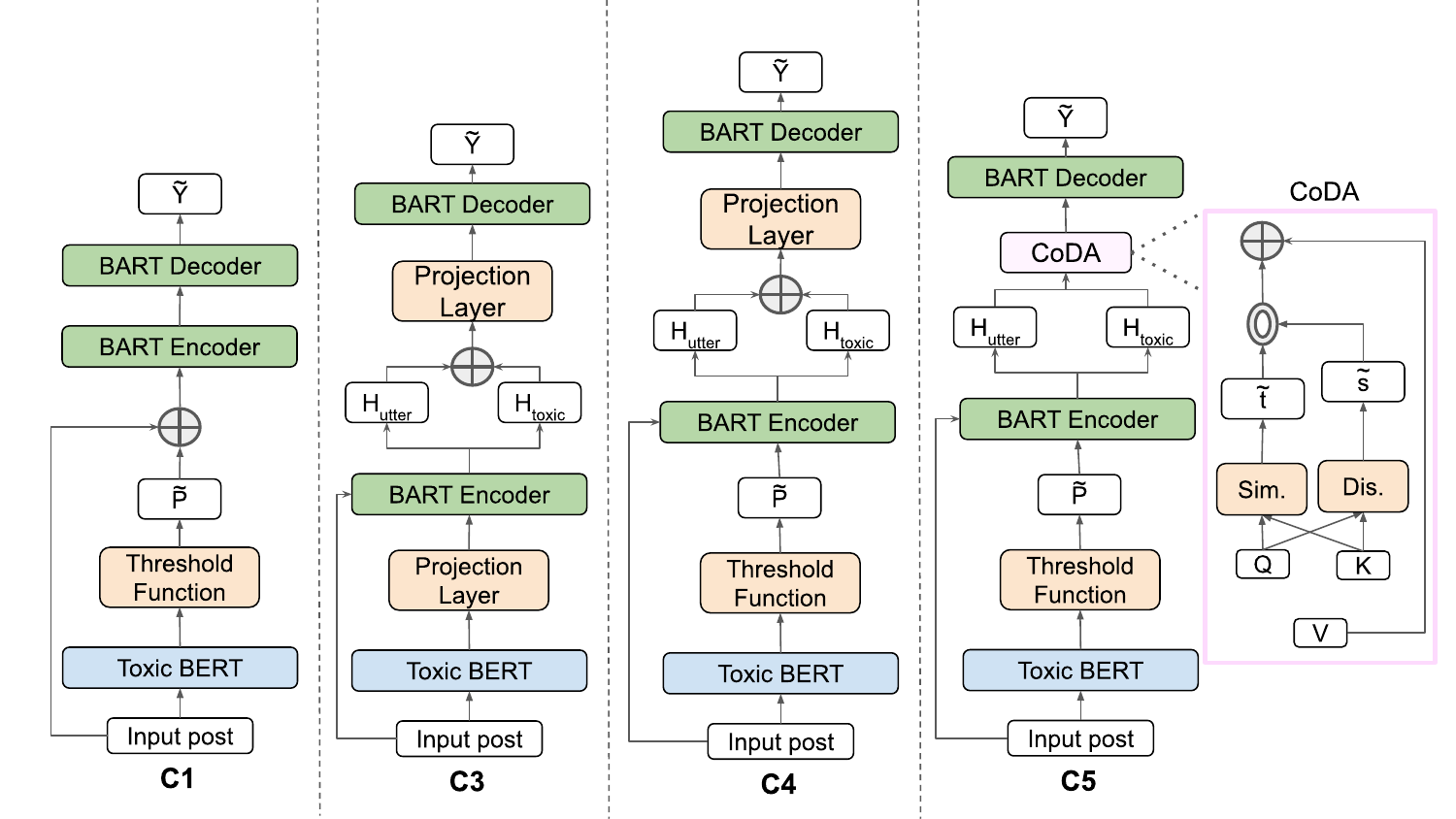}
\caption{Configurations (C1, C3-C5) for incorporating in-domain toxic attributes obtained from the ToxicBERT regressor. The toxic-attributed BART so finetuned is our proposed system - \toxbart. BART encoded toxic attributes, and input representations are H\textsubscript{toxic} and H\textsubscript{utter}, respectively. $\mathcal{\tilde{P}}$ is the modified toxic attributes vector, whereas $\tilde{Y}$ is the system generated explanations. For the CoDA setup, query (Q), Key (k), and value (V) interactions are captured by Sim. (similarity) and Dis. (dissimilarity) matrices. Here,  $\tilde{t}$ and  $\tilde{s}$ represent the tanh and sigmoid functions, respectively. 
}
  \label{fig:model_config}
\end{SCfigure*}

\section{Related Work}
One way to help moderators is to incorporate the context by uncovering stereotypical implications. The line of work involving explaining toxic text \cite{cao-etal-2022-intrinsic, balkir-etal-2022-necessity} or generating stereotypical implications \cite{sridhar-yang-2022-explaining, sap-etal-2020-social, elsherief-etal-2021-latent} is nascent and primarily employs variants of large language models (LLMs) \cite{zhou-etal-2023-cobra, mun-etal-2023-beyond, zhang-etal-2023-biasx}. Meanwhile, post hoc attention scoring and rationale-based training techniques \cite{hatexplain-aaai21,10.1145/3534678.3539161} fail to detect implicit spans.

\textbf{In-context Learning.} LLMs can perform complex tasks with the help of demonstrations \cite{liu-etal-2022-makes} via in-context learning (ICL) \cite{gpt3-paper-brown} and prompt engineering \cite{singhal2022large}. However, limitations of exemplars having negligible effects on the LLM performance \cite{min-etal-2022-rethinking, liu-2022-few} or on out-of-domain samples have also been reported \cite{an2022inputtuning,lyu-etal-2023-z}. The role of demonstrations and prompting has been examined in hate speech as well \cite{10.1145/3543873.3587320,yang2023hare}. Our observations for infusing KG tuples are similar to employing examples \cite{min-etal-2022-rethinking,lin-2022-leveraging}, but differ in that we examine generative tasks. 

\textbf{Leveraging External Knowledge.} Knowledge graphs are often applied in NLP \cite{schneider-etal-2022-decade,10387715,10.1145/3512467}. Leveraging commonsense KGs \cite{conceptnet5-speer, sap-atomic-2018} has been explored for reasoning \cite{chang-etal-2020-incorporating}, question answering \cite{feng-etal-2020-scalable}, story generation \cite{10.1162/tacl_a_00302}, sarcasm explanation \cite{kumar-etal-2022-become}, etc. The role of knowledge graphs and world knowledge (Wikipedia summaries) have also been explored for hate target detection \cite{ReyeroLobo2023}, and implicit type classification \cite{elsherief-etal-2021-latent,lin-2022-leveraging}. Meanwhile, \citet{deshpande-etal-2022-stereokg} released a stereotype-focused KG targeting six nationalities and religions. 

Our work builds upon a BART-based solution, MIXGEN \cite{sridhar-yang-2022-explaining}. MIXGEN is an ensemble of three different knowledge signals (\emph{expert}, \emph{implicit}, \emph{explicit}) that generate implied explanations. Their definition of \emph{expert} knowledge\footnote{During the examination, we were not able to obtain these expert attributes and were able to reproduce only two of their modules, i.e., the \emph{explicit} and \emph{implicit}.} is similar to our `in-dataset` toxicity signal. The \emph{expert} knowledge was obtained as extra annotations for the dataset. MIXGEN utilized \emph{explicit} knowledge in the form of top-k \concept\ tuples. The entities and relations ``explicitly" mentioned via the top-k KG tuples should nudge the PLM to focus on relevant aspects of the input. The \emph{implicit} knowledge was obtained via prompted outputs from GPT-2. The contextual signals from a language model (LM) are `implicit' as these are nudged from within the latent space of the LM, having access to world knowledge through its training.

\section{Infusing Toxicity Attributes for Explaining Implicit Hate}
\label{sec:tox_bart}
We first outline the in-domain ($\mathcal{P}$) and in-dataset ($\mathcal{A}$) ``toxicity attributes'' and then formulate multiple configurations to incorporate them with BART. The \toxbartfull\ (\toxbart) is then tuned on a pair of implicit input posts ($\mathcal{X}$) and implied explanations ($\mathcal{Y}$). We denote the BART encoder/decoder with $\mathcal{F}_\theta$/$\mathcal{G_\theta}$, with $d \in \mathbb{R}^{768}$ embedding dimension and $\theta$ trainable parameters.

\noindent \textbf{In-domain Attributes.} These are external to the dataset but related to the ``domain'' of hate speech, conveying information about the harmfulness of the incoming posts. Here, we employ the large-scale Jigsaw toxicity dataset ($\approx 2M$ datapoints) \cite{jigsaw-unintended-bias-in-toxicity-classification} to facilitate the same. In Jigsaw, an input text $j$ has multiple annotations, with each annotator giving a score between $0-1$ for labels $t_1,t_2,\ldots,t_6 \in$ \{toxicity, severe toxicity, obscene, threat, insult, identity attack\}. We leverage these scores by training a BERT regressor with $6$ regression heads (one for each label). 

Formally, given a regressor $\mathcal{R}_{\phi}$ with parameters $\phi$, input $j$, and labels $t$, we minimize $\mathcal{L}_{MSE} = \frac{1}{n}\sum_{i=1}^{n}(\mathcal{R}_{\phi}(j_i) - t_i)^2$. The train-test RMSE of \texttt{ToxicBERT} is enlisted in Appendix \ref{app:toxbert}. The output of \texttt{ToxicBERT} is a vector $\mathcal{P} \in \mathbb{R}^{1*6}$, where each dimension represents the probability for the type of insult. We explore different configurations to infuse $\mathcal{P}$ with the incoming post $\mathcal{X}$ (Figure \ref{fig:model_config}). Below, we expand on the best configurations observed for infusing \textit{in-domain} attributes. Appendix \ref{app:configs} outlines the rest of the configurations.

\textbf{Configuration 1 (C1).} The probability values in isolation do not convey information about the toxicity attributes. Hence, we convert the values into their corresponding toxicity tokens via a threshold parameter ($\lambda$). For instance, if $p_i$ captures the probability score for the label ``threat," then based on  $p_i<\lambda$, its equivalent textual presentation will be a special token either $<$NOT\_THREAT$>$ or $<$THREAT$>$. The six toxicity tokens are then concatenated (using $[SEP]$) to the incoming posts ($\mathcal{X}$). Employing tokens provides more uniformity, as the chances of token sets to co-occur are higher than that of exact probability score vectors. Equation \ref{eq:c2} outlines the setup, where $\Gamma$ corresponds to the probability scores to toxicity-token transformation function parametrized by $\lambda$.
\begin{equation}
  \begin{gathered}
    \mathcal{\tilde{P}} = \Gamma(\lambda, \mathcal{P}) \\
    \quad \tilde{X} = [\mathcal{X}, \mathcal{\tilde{P}}] ; \quad \mathcal{\tilde{Y}} = \mathcal{G_\theta}(\mathcal{F_\theta}(\tilde{X}))
\end{gathered}
\label{eq:c2}
\end{equation}

\noindent \textbf{In-dataset Attributes.} These are supplementary annotations already available within the respective dataset. For example, both \sbic\ and \latent\ have free-text annotations for the target group. \sbic\ further has labels indicating whether the incoming posts are (a) \textit{intentional}, (b) \textit{lewd}, (c) \textit{offensive}, (d) \textit{targeting a group}, and (e) \textit{uses in-group language}. Meanwhile, \latent\ has labels indicating the type of implicit hate from among -- {\textit{grievance, incitement, inferiority, irony, stereotypical, threatening}, or \textit{other}. 

\textbf{Configuration 2 (C2):} For $n$ ``in-dataset'' attributes $A=\{A_1, A_2 \ldots \, A_n\}$ for an input post, we first concatenate them using whitespace ($\tilde{A} = [A_1[w]A_2..\ldots [w]A_n]$) and then concatinate $\tilde{A}$ with input post as outlined in  Equations \ref{eq:c5}.
\begin{equation}
        \tilde{X}=[X,\tilde{A}] ;
        \mathcal{\tilde{Y}} = \mathcal{G_\theta}(\mathcal{F_\theta}(\tilde{X}))
\label{eq:c5}
\end{equation}

\noindent\textbf{Overall Loss.} For every configuration, we aim to reduce the cross-entropy loss over the predicted generations $\mathcal{\tilde{Y}}$ infused by toxicity attributes ($\mathcal{P}$ or $\mathcal{A}$) in \toxbart\ based on $\mathcal{L}_{CE} = \frac{1}{m}\sum_{i=1}^{m}(\mathcal{Y},\mathcal{\tilde{Y}})$.

\begin{table}[!t]
\resizebox{\columnwidth}{!}{
    \begin{tabular}{c|cc|cc}
    \hline
    \textbf{Feature}   & \multicolumn{2}{c|}{\textbf{\sbic}} & \multicolumn{2}{c}{\textbf{\latent}}\\ \cline{2-5}
     & \textbf{Train} & \textbf{Test} & \textbf{Train} & \textbf{Test} \\\hline
        \# Samples & 35933 & 4705 & 5722 & 636 \\
        Post len. & 107.0 (63.3) & 107.0 (65.6) & 94.0 (40.0) & 31.0 (11.7)\\
        Implied len.  & 16.0 (15.3) & 19.0 (14.5) & 96.0 (43.8) & 31.0 (11.7) \\ \hline
    \end{tabular}}
    \caption{Dataset statistics enlisting the number of train and test samples in \sbic\ and \latent. Here, `post' is the input implicit statement, and `implied' is the implied stereotype. We report both features' average (standard deviation) token length (len).}
    \label{tab:dataset_statistics}
\end{table}

\section{Impact of Infusing Toxicity Attributes}
\label{sec:exp_rq2}
To establish the efficacy of ``toxicity attributes,'' we conduct extensive automatic and human evaluation comparing \toxbart\ with KG and non-KG-based systems. Further, we show the robustness and sensitivity of \toxbart via ablation. The experimental setup is enlisted in Appendix \ref{appendix:exp_details}.

\textbf{Data Source.} We employ \sbic\ \cite{sap-etal-2020-social} and \latent\ \cite{elsherief-etal-2021-latent} datasets containing $\approx35k$ and $\approx4k$ samples respectively. Both are a parallel corpus of an input post obtained from the web containing implicit hate ($\mathcal{X}$) and the corresponding stereotype explanation ($\mathcal{Y}$) obtained via human annotations. A single post from \sbic\ can have multiple annotations. For \latent\, every post has a single annotation. The dataset statistics of \sbic\ and \latent\ are enlisted in Table \ref{tab:dataset_statistics}. Hateful posts for \sbic\ \cite{sap-etal-2020-social} are sourced in equal parts from Reddit, Twitter, and ExtremeHate Forums (Gab, Stormfront, BannedReddit). Meanwhile, \latent\ \cite{elsherief-etal-2021-latent} is solely curated from Twitter. \concept\ \cite{conceptnet5-speer} is a KG consisting of $\approx 34M$ tuples/assertions of world knowledge and common sense relations curated from Wikipedia.

\textbf{Baseline Systems.} We start with vanilla PLMs (BART and GPT-2) finetuned without any external attribute. We then access external attributes via MIXGEN's\footnote{We observe a significant deviation in results reproduced for MIXGEN. Since we did not change or tune any hyperparameters from the original MIXGEN setup during training and inference, this discrepancy can arise from hardware or random seeding currently missing from MIXGEN.} \emph{explicit knowledge} and \emph{implicit knowledge} signals. Finally, we compare the \emph{zero-shot} generations of GPT-3.5-Turbo}. We employ the following prompt for generating implications \textit{``What stereotype is propagated by this post: [POST]? Answer in simple words and keep the length short''}. As this study aims to focus on smaller-grade fine-tunable PLMs, we do not perform extensive prompt engineering for GPT-3.5. However, after the initial investigation, we added the phrase ``answer in simple words and keep the length short'' to reduce wordy\footnote{Based on dataset statistics in Table \ref{tab:dataset_statistics} mean explanation length is $\approx 25$ words.} and non-contextual explanations like ``People should not indulge in hateful content.''

\begin{table}[t]
\centering
\resizebox{\columnwidth}{!}{
\begin{tabular}{lcccccccc}
\toprule
    \multirow{2}{*}{\textbf{Method}} & \multicolumn{3}{c}{\bf \sbic} & & \multicolumn{3}{c}{\bf \latent}\\
    \cmidrule{2-4} \cmidrule{6-8}
    & \textbf{B} & \textbf{R} & \textbf{BS} & & \textbf{B} & \textbf{R} & \textbf{BS}\\
    \midrule
    GPT-2 & 62.72 & 62.72 & 59.04 &  & 30.94 & 21.99 & 82.71 \\
    BART & \textbf{72.17} & \textbf{70.83} & 78.05 & & 38.38 & 17.65 & 90.37 \\ \hdashline
    MIXGEN - \textit{Imp} & \underline{72.12} & \underline{69.84} & \underline{80.91} & & 46.28 & 35.78 & 92.09 \\
    MIXGEN - \textit{Exp} & 68.41 & 66.40 & 80.37 & & \underline{47.23} & \textbf{36.26} & \underline{92.12} \\
    MIXGEN - \textit{Exp} + \textit{Imp} & 70.27 & 67.69 & 80.23 & & 47.00 & 33.09 & 90.8 \\ \hdashline
    \toxbart\textsubscript{C1} & 64.89 & 63.83 & 64.52 & & 41.94 & 26.28 & 89.47 \\
    \toxbart\textsubscript{C2} & 69.85 & 68.23 & 75.78 & & \textbf{47.72} & \underline{34.70} & \textbf{92.89} \\ \hdashline
    GPT-3.5 (Zeroshot) & 37.45 & 15.36 & \textbf{90.10} & & 33.57 & 10.40 & 90.06 \\ \bottomrule
\end{tabular}}
\caption{Results for generating explainations for implicit stereotypes for \sbic\ and \latent. Bold (underlined) values represent the best-performing (second-best) setup for the given dataset for -- B: max-BLEU; R: ROUGE-L F1; BS: BERTScore F1. For MIXGEN's implicit (explicit) signal infusion, we keep $k_i=15$ ($k=20$) as adopted from \citet{sridhar-yang-2022-explaining}. }
\label{table:final_results}
\end{table}

\begin{table*}
\subfloat[]{
\resizebox{0.725\textwidth}{!}{
    \begin{tabular}{lccccc}
    \hline
        \textbf{Method} & \textbf{Flu.} & \textbf{Coh.} & \textbf{Spe.} & \textbf{Sim.} & \textbf{Tar.}\\
        \toxbart\textsubscript{C1} & 4.52 ($\pm 0.76$) & 3.95 ($\pm 0.99$) & 3.67 ($\pm 0.92$) & 3.47 ($\pm 1.00$) & 0.78 ($\pm 0.29$) \\
        GPT-3.5 & 4.17 ($\pm 0.9$) & 3.74 ($\pm 0.92$) & 3.27 ($\pm 1.07$) & 2.78 ($\pm 1.14$) & 0.49 ($\pm 0.4$)\\\hline
    \end{tabular}
    }
    }
\subfloat[]{
\resizebox{0.25\textwidth}{!}{
    \begin{tabular}{lc}
    \hline
        \textbf{Method} & \textbf{Toxicity}$\uparrow$\\
        \toxbart\textsubscript{C1} & 0.89 ($\pm 0.21$) \\
        GPT-3.5 & 0.33 ($\pm 0.32$)\\\hline
    \end{tabular}
    }
    }
\caption{Comparison between \toxbart\textsubscript{C1} and GPT-3.5 on $17$ random samples of \sbic\ via -- (a) Mean (std) score for each metric from $20$ evaluators. Fluency (Flu.), coherence (Coh.), specificity (Spe.), and gold similarity (Sim.) are on a 5-point scale, and the target detection  (Tar.) is boolean. (b) Toxicity scores from Unitary toxicity API. The higher the toxicity, the closer to the intended explicit connotation of the explanations.}
\label{tab:hum_eval}
\end{table*}

\begin{table*}[!t]
\centering
\resizebox{\textwidth}{!}{
\begin{tabular}{p{14em}|p{9em}|p{9em}|p{20em}}
\hline
 \textbf{Utterance} & \textbf{\toxbart\textsubscript{C1}} & \textbf{GPT-3.5 (zero-shot)} & \textbf{Gold Label Stereotype(s)} \\ \hline
You used to be able to walk into a shop with £2.00 and leave with all you could possibly need. Now all the blacks are beating us to it. & black people are thieves & Racist stereotype about black people. & black people steal, blacks are getting there before whites, there are too many black people, blacks are beating whites to the punch, black people are greedy, black people are always taking things\\ \hdashline
What’s the difference between science and religion? Science will fly you to mars. Religion will fly you into a skyscraper & muslims are terrorists. & Science = Mars, Religion = Skyscraper. & islamic folks religion is only about hurting people, religion is bad, islamic folks religion is a joke, religion makes you stupid \\ \hdashline
Why are all women left handed? Because they don't have rights. & women don't have rights & Women are not capable. & women are slaves, women are inferior, don't have all rights men have, women are stupid, are missing rights that they should have, women have no rights \\
\bottomrule
\end{tabular}}
\caption{Some samples taken (verbatim) from \sbic\ and their predicted explanations by \toxbart\textsubscript{C1} and GPT-3.5 describing the clear difference between the quality of generations.}
\label{tab:our_decent_predictions}
\end{table*}

\paragraph{\bf Automated Evaluation.} We employ BLEU \cite{papineni-etal-2002-bleu}, ROUGE-L \cite{lin-2004-rouge}, and BERTScore \cite{tianyi-bertscore} to measure the syntactic, linguistic, and semantic similarities between the generations and gold labels. Interestingly, for \sbic\, infusion of external signal (KG or toxicity) leads to a drop in its performance compared to vanilla BART. On the other hand, the \latent\ dataset proves more difficult with fewer samples to train on; however, infusion of external signals leads to performance improvement over vanilla BART. For both \sbic\ and \latent\ \textit{in-dataset} attributes (\toxbart\textsubscript{C2}) perform at par with MIXGEN. The comparatively better performance of \textit{in-dataset} against \textit{in-domain} features reinstates the importance of human-in-the-loop to mitigate hatefulness.  

For \sbic, \toxbart\textsubscript{C2} displays comparable performance to both the MIXGEN setups -- implicit and explicit knowledge with only a slight variation of (-$2.27$, -$1.61$, -$5.13$) and (+$0.44$, +$1.83$, -$4.59$) points in (BLEU, ROUGE-L, and BERTScore). In \latent, \toxbart\textsubscript{C2} perform at par with MIXGEN with (BLEU, ROUGE-L, and BERTScore) scores of ($1.44$, $-1.08$, $0.8$) and ($0.49$, $-1.56$, $0.77$) for implicit and explicit knowledge baselines. \toxbart\textsubscript{C2} beats the vanilla BART by ($9.34$, $17.05$, $2.52$) points in (BLEU, ROUGE-L and BERTScore). 

Table \ref{table:final_results} also highlights that based on standard lexical metrics, zero-shot systems underperform finetuned PLMs. However, GPT-3.5 produces higher semantic scores ($>90$ BERTScores). We hypothesize this discrepancy in lexical metric arises as the train-test distribution for our finetuned PLMs is closer than the zero-shot setup for GPT-3.5. We perform a human evaluation to assess further the semantic richness of \toxbart\ and GPT-3.5. 

\paragraph{\bf Human Evaluation.} It is performed between \toxbart\textsubscript{C1} vs. GPT-3.5, assuming the evaluators are proxies for content moderators. They are provided anonymized outputs from both systems against a given input sample, gold generation, and a gold target label. $20$ evaluators access $17$ random samples from \sbic\ on $5$ metrics -- \emph{Fluency, Coherency, Specificity, Similarity with gold explanation, and Target Group}. \emph{Fluency} and \emph{Coherency} measure the broader grammatical correctness.  \emph{Specificity}, \emph{Similarity with gold explanation}, and \emph{Target Group} capture the task-specific correctness of how well the model presents the underlying stereotype. Appendix \ref{app:human_eval} lists the details of human evaluation.

A manual analysis (Table \ref{tab:our_decent_predictions}) of the GPT-3.5-based generation reveals its tendency to produce non-specific/broad-stroke explanations. It may stem from GPT-3.5 being trained/filtered to discourage harmful discourse. Our investigation aligns with practitioners' observations that GPT-3.5-based LLMs are rigorously guardrailed, hampering their ability to perform well in tasks such as ours, necessitating the generation to be explicit and specific about stereotypes. For example, looking at the first instance in Table \ref{tab:our_decent_predictions}, we see that the terms ``Racist stereotype'' and ``black people'' are semantically close to the gold generations, even though it is not specific. Since BertScore \cite{tianyi-bertscore} employs pair-wise semantic embedding matching, using generic terms that are semantically closer to the target group helps GPT-3.5 maintain the high BertScore. Yet, it leads to higher variability on \textit{Specificity} (Table \ref{tab:hum_eval} (a)) for GPT-3.5.
 
We further corroborate the generality of the explanations from GPT-3.5 by computing toxicity scores from Unitary toxicity API \cite{detoxify-unitary}. On average, \toxbart\textsubscript{C1}'s  generations are much more toxic compared to GPT-3.5 ($0.89$ vs. $0.33$), as observed in Table \ref{tab:hum_eval} (b). \emph{As we aim to unmask the underlying stereotype, the generated output is expected to be explicit.} 

\emph{Our system is intended to help content moderators. The more straightforward and explicit (and therefore seemingly toxic) the explanations, the better the content moderators will be to judge the incoming implicit hate. It is important to reiterate that an increase in explicitness comes at the cost of specificity. We observe that \toxbart\ can achieve optimal performance in balancing the explicitness while retaining the specificity of the target group and the underlying stereotype as supported by automated (Table \ref{table:final_results}) and human evaluations (Tables \ref{tab:hum_eval} and \ref{tab:our_decent_predictions}). Our evaluations, thereby, point towards \toxbart\ achieving the intended usage as highlighted by the initial motivation in Figure \ref{fig:motivation}.}

\begin{table}[t]
    \centering
    \resizebox{\columnwidth}{!}{
    \begin{tabular}{lccc}
        \hline
        \textbf{Method} & \textbf{BLEU} & \textbf{ROUGE-L} & \textbf{BERTScore} \\\hline
        \toxbart\textsubscript{C1} & 64.89 & 63.83 & 64.52 \\ \hdashline
        \toxbart\textsubscript{C1}\textsuperscript{Exp. 1} & 68.92 & 67.16 & 72.32 \\
        \toxbart\textsubscript{C1}\textsuperscript{Exp. 2} & 63.74 & 63.47 & 61.21 \\
        \toxbart\textsubscript{C1}\textsuperscript{Exp. 3} & 62.8 & 62.76 & 59.34 \\\hdashline
        \toxbart\textsubscript{C1}\textsuperscript{Exp. 4a} & 63.16 & 62.96 & 60.11 \\
        \toxbart\textsubscript{C1}\textsuperscript{Exp. 4b} & 62.95 & 62.82 & 59.72 \\
        \toxbart\textsubscript{C1}\textsuperscript{Exp. 4c} & 64.73 & 64.25 & 62.79 \\\hdashline
        \toxbart\textsubscript{C1}\textsuperscript{Exp. 5a} & 65.16 & 64.24 & 64.87 \\
        \toxbart\textsubscript{C1}\textsuperscript{Exp. 5b} & 74.64 & 72.45 & 83.89 \\
        \toxbart\textsubscript{C1}\textsuperscript{Exp. 5c} & 63.55 & 63.36 & 60.63 \\
        \toxbart\textsubscript{C1}\textsuperscript{Exp. 5d} & 63.9 & 63.45 & 61.85 \\\hline
    \end{tabular}}
    \caption{Ablations on \toxbart\textsubscript{C1} on \sbic. The first set adjusts hyperparameters of C1 (Exp 1-3). The second set perturbs toxicity probabilities (Exp 4 a-c). The final set flips the attribute label (Exp 5 a-d).}
    \label{tab:ablation_exp}
\end{table}

\begin{table*}[t]
\centering
\resizebox{\textwidth}{!}{
\begin{tabular}{c|p{18em}|p{11em}|p{7em}|p{10em}|p{10em}}
\hline
\textbf{\#} & \textbf{Utterance} & \textbf{Implied Stereotype} & \textbf{Target Group} & \textbf{Stereotype Prediction} & \textbf{Error Type} \\ \hline
1 & ``Oh, you mean you're Chinese, right?" & `' & `' & asian folks are not smart & Hallucinating Stereotype \\ \hdashline
2 & How's America like the Dark Knight Trilogy? No one gives a fuck about Harvey. & trivializes harm to victims &  climate disaster victims & `' & Incomplete Stereotype \\ \hdashline
3 & what do you call a nine year old with no friends? a sandy hook survivor & are losers.; everyone else is dead; school shooters are always socially isolated. & mass shooting victims & children are targets of pedos & Wrong Target Group + Incorrect/Incomplete Annotation \\ \hdashline
4 & my friend thought she was lesbian just for having short hair. $<$repeat$>$ ``you can't assume that," i said, ``she's not obese." &  lesbian women are fat; lesbian women are unattractive & lesbian women & lesbians have short hair & Subjective annotations \\
\bottomrule
\end{tabular}}
\caption{Some samples taken (verbatim) from \sbic\ and the predicted explanations by \toxbart\textsubscript{C1} describing the error types. We also highlight if the predicted model captures the target correctly for the gold target group. Empty strings (`') correspond to either no gold label or no generation by the model.}
\label{tab:error_analysis_samples}
\end{table*}

\paragraph{\bf Ablation Study.} We perform ablations on our ``in-domain" attributed setup (\toxbart\textsubscript{C1}) using \sbic. In the first set of experiments, we alter \toxbart\textsubscript{C1} under various settings. In the first setting (\textbf{Exp. 1}), keeping all hyperparameters the same, we replace the toxicity tokens with pre-defined plain text, which is not a special token as provided in Table \ref{tab:ablation_experiment_prompts}. From Table \ref{tab:ablation_exp}, we observe that a pre-defined prompt token as a feature significantly improves the performance. In the second (\textbf{Exp. 2}) and third (\textbf{Exp. 3}) settings, we vary the threshold $\lambda = \{0.3,0.6\}$ on \toxbart\textsubscript{C1}. Among $\lambda = \{0.3,0.6,0.5\}$, though the difference is small, the default $\lambda = 0.5$ works best.



\paragraph{\bf Impact of Toxicity Probabilities.} To measure the flexibility of these attributes, we also perform an experiment by perturbing the input probability scores (\textbf{Exp. 4}). None of these settings obtain scores from the trained BERT regressor but rather generate them by making our toxicity attribute probabilities as: a) all zeros, b) all ones, and c) random (between 0-1). As observed from Table \ref{tab:ablation_exp} \emph{the three adversarial configurations from \textbf{Exp. 4} register an expected deterioration in performance.} These observations strengthen our initial decision to opt for toxicity attribute infusion.

\paragraph{\bf Impact of flipping Toxic Attributes.} We also measure the sensitivity of the model w.r.t flipped attributes (replacing $A_{i}$ by its counterpart $\neg A_{i}$). The results for the same are illustrated in \textbf{Exp. 5} of the Table \ref{tab:ablation_exp}. We perform these experiments for the top 4 attributes -- toxic (\textbf{Exp. 5a}), severely toxic (\textbf{Exp. 5b}), obscene (\textbf{Exp. 5c}) and threat (\textbf{Exp. 5d}) with the lowest occurrence rates of attributes with a probability greater than the threshold. We make an intriguing observation where flipping the severe toxic labels caused the model's performance to overshoot well beyond the baselines. Since explaining implicit stereotypes aims to bring out the explicitness of a statement, we observe that highlighting an incoming post as extremely toxic nudges the model to produce more explicit explanations. However, this only occurs when employing severely toxic or toxic attributes. \emph{This uncanny observation calls into question the need for interoperability studies on how augmentation of external signals nudge generations. We hypothesize that domain-specific generative language models are susceptible to extreme attributes from the same domain.}

\begin{table*}[!t]
\resizebox{\textwidth}{!}{
\begin{tabular}{lccc|ccc|ccc|ccc}
\hline
     \multirow{3}{*}{\textbf{Method}} & \multicolumn{6}{c}{\textbf{\concept}} & \multicolumn{6}{|c}{\bf \texttt{StereoKG}} \\ \cline{2-13}
     & \multicolumn{3}{c}{\bf \sbic} & \multicolumn{3}{c}{\bf \latent} & \multicolumn{3}{|c}{\bf \sbic} & \multicolumn{3}{c}{\bf \latent} \\ \cline{2-13}
     & \textbf{B} &	\textbf{R} & \textbf{BS} & \textbf{B} &	\textbf{R} & \textbf{BS} & \textbf{B} &	\textbf{R} & \textbf{BS} & \textbf{B} &	\textbf{R} & \textbf{BS} \\\hline
    BART Baseline & 72.17 & 70.83 & 78.05 & 38.38 & 17.65 & 90.37 & 72.17 & 70.83 & 78.05 & 38.38 & 17.65 & 90.37 \\ \hdashline
    Top-k & 68.41 & 66.4 & 80.37 & 47.23 & 36.26 & 92.12 & 63.57 & 61.30 & 76.39 & 46.39 & 35.37 & 92.03 \\
    Bottom-k & 68.97 & 66.80 & 80.95 & 47.40 & 35.90 & 92.15 & 60.31 & 58.09 & 73.44 & 46.92 & 35.94 & 92.04 \\
    Random-k & 69.69 & 67.47 & 81.63 & 48.34 & 37.18 & 92.31 & 60.80 & 58.45 & 73.87 & 47.27 & 36.12 & 92.07 \\
    \hline
\end{tabular}
}
\caption{\sbic\ and \latent's performance variation across \concept\ and \stereo\ in terms of -- B: max-BLEU; R: ROUGE-L F1; BS: BERTScore F1. The KG-tuples ($k$) are concatenated with BART input tokens to generate explanations. $k=20$ is the best-performing hyperparameter of MIXGEN \cite{sridhar-yang-2022-explaining}.}
\label{table:quality_cos_results}
\end{table*}

\paragraph{\bf Error Analysis.} Here, we broadly discuss two classes of errors via \toxbart\textsubscript{C1} on \sbic. 

\noindent $\bullet$ \ul{Modeling Errors:} While training \toxbart, we observe that the \sbic\ dataset has some empty rows (aka no gold explanations). For example, case \#$1$ in Table \ref{tab:error_analysis_samples} is hard to annotate without knowing if the question is out of curiosity, sarcasm, or disdain. Despite this, \toxbart\ and even other baselines generate implied stereotypes, leading to ``hallucinated'' explanations. Meanwhile, there were cases where the model failed to generate contextual explanation, as highlighted in case \#$2$ in Table \ref{tab:error_analysis_samples}. In  \#$2$, \toxbart\ misses the climate reference. Lastly, we observe that in some instances, the LLM misidentifies the target and the subsequent explanation. For example, the focus on ``nine-year-old'' in case \#$3$.

\noindent $\bullet$ \ul{Annotation Errors:} While performing the preprocessing and manual evaluation of predictions, we notice that both \sbic\ and \latent\ have mislabeling, leading to incorrect gold explanations. For example, in case \#$3$, some annotators provide incomplete sentences like ``are losers'' or phrases that can be triggering for the target group like ``everyone else is dead'' w.r.t school shooting. We also note that annotations can be highly subjective. For case \#$4$ in Table \ref{tab:error_analysis_samples}, multiple stereotypes are true, each based on the annotator's knowledge and prejudice. In this case, the predicted stereotype, though valid, is not covered in the ground annotation set. 

\section{Auditing the quality of KG tuples}
\label{section:cos_quality_relevance}
While establishing \toxbart's efficacy in Section \ref{sec:exp_rq2}, we also observe that MIXGEN's ``explicit knowledge'' augmented via \concept leads to a drop in performance for \sbic, while providing a marginal improvement on \latent. Given the prevalence of KG augmentation in NLP \cite{schneider-etal-2022-decade}, we are motivated to establish a relation between the ``quality'' of knowledge tuples and the generations for stereotype explanation. 

\textbf{Setup.} Directly establishing the causal relation between the quality of KG tuples and the generated output from BART is intractable. Instead, we hypothesize that: \textit{if adding top-k KG helps improve a model's generation capabilities, then the generations should deteriorate when the top-k is corrupted.} We investigate this via KG-infusion for BART on \sbic\ and \latent. To better understand the role of KG, we employ two conceptually different KGs. \concept\ is a large-scale KG curated from Wikipedia. \stereo\ \cite{deshpande-etal-2022-stereokg} is a nascent KG with $4k$ tuples capturing stereotypes from Twitter and Reddit. Given the intention of capturing stereotypes in social media posts, \stereo\ is closest to being an ideal KG for our task. An overview of the KGs is provided in Table \ref{tab:kg_prop} (Appendix \ref{app:kg_exp}).

We concatenate the input post ($\mathcal{X}$) with $k$ tuples ($t_1,t_2,\ldots,t_k$) as $\mathcal{\tilde{X}}$ = \texttt{\{$\mathcal{X}$, $[SEP]$, $t_1$, $[SEP]$, $t_2$, $[SEP]$,\ldots, $t_k$\}}, where $[SEP]$ is the separator token. $\mathcal{\tilde{X}}$  is then input to BART. The outline of how the $k$ tuples are retrieved from respective KG is provided in Appendix \ref{app:kg_exp}. 

\textbf{Observations.}
Table \ref{table:quality_cos_results} shows that compared to standalone BART, \latent's performance improves under all KG infusion. Meanwhile, due to KG infusion, \sbic\ is more varied and even registers a drop in BLEU and ROUGE-L. More interestingly, we have counter-intuitive results comparing the three top/bottom/random-$k$ configurations. In 3/4 combinations, \emph{the performance difference is visibly insignificant (and in some instances even increases) if we replace top-k with bottom-k or random-k tuples.} \emph{While the influence of KG on a dataset varies on a case-by-case basis, there is a noticeable deviation in expected behavior for incorporating bottom and random-k tuples.}

 \begin{table}[!t]
\resizebox{\columnwidth}{!}{
\begin{tabular}{l|l|l|l|l}
\hline
\textbf{KG}          & & \textbf{Base}  & \textbf{T}                             & \textbf{B}     \\\hline
\multirow{3}{*}{C} & \textbf{T}                             & 2.19**, 2.56**, 1.74**  &                                            & \\
& \textbf{B}                          & 2.06**, 2.23**, 1.89**  & 0.25, -0.01, -0.15  & \\
& \textbf{R}  & 2.02**, 2.18**, 1.46** & 0.28, -0.21, -0.28 &  -0.00, -0.22, -0.17 \\ \hdashline                          
\multirow{3}{*}{S} & \textbf{T}                            & 2.21**, 2.00**, 1.33** &                                            & \\
& \textbf{B}                          & 2.12**, 2.06**, 1.71**  & 0.37, 0.24, 0.33	  & \\
& \textbf{R}  & 2.24**, 2.49**, 1.42** & 0.25, 0.08, -0.16 &  -0.09, -0.13, -0.40* \\ \hline                                      
\end{tabular}
}
\caption{Pair-wise Effect size and p-test on (B: max-BLEU; R: ROUGE-L F1; BS: BERTScore F1) when comparing the column-wise control group with the row-wise treatment group for \latent\ on BART-base with \concept\ and \stereo\ respectively, with $k=20$. * ($p \le 0.05$) and **($p \le 0.001$) indicate whether the difference in pairwise metric is significant.}
\label{tab:kg_ptest}
\end{table}

\textbf{Hypothesis testing of KG influence.} Given the higher deviation in performance for \latent, we also report the paired t-test and each pair's effect size under consideration on \latent. We report variation in all three metrics. Based on Table \ref{tab:kg_ptest}, we see that going from vanilla BART to KG infusion (top, bottom, or random) leads to a significant increase in performance, as corroborated by a considerable effect size ($\geq1$) and $p\leq0.01$ in all metrics for the ``Base'' column in both \stereo\ and \concept. On the other hand, among top-k, bottom-k, and random-k, the small effect sizes effectively capture a slight increase or decrease in performance metrics in Table \ref{table:quality_cos_results}. Here, the insignificant ($p>0.01$) effect size of small negative values indicates that the considerably negligible variation among top, bottom, and random-k can be by chance and that replacing one with the will not significantly alter the performance.

\textbf{Retrieval scores.} The range for retrieval scores termed as relevance and similarity scores, respectively, for \concept\ and \stereo\ is  [$0,\inf$) and [$0,1$] (details in Appendix \ref{app:kg_exp}). Figure \ref{fig:kg_error} shows that patterns of scores per KG are similar for respective hate datasets, proportional to the number of test samples in each. The majority of relevance scores w.r.t \concept\ are $\leq1$ and only $3.5\%$ ($1.5\%$) of samples of \sbic\ (\latent) garner scores $>=5$ for at least one of the tuple. The similarity scores for \stereo\ are also on the lower end, with the majority covered in the range $0.3-0.5$. \emph{These observations indicate low-quality tuples getting filtered in top-k.} Based on top-k retrieval scores, bottom-k and random-k should be equally low-quality. In Appendix \ref{app:kg_exp}, we also look at the uniqueness of retrieval scores.

\begin{figure}[!t]
\resizebox{\columnwidth}{!}{
\begin{tabular}{@{}cc@{}}
    \subfloat[\concept]{%
        \label{c}%
        \includegraphics[width=0.28\textwidth]{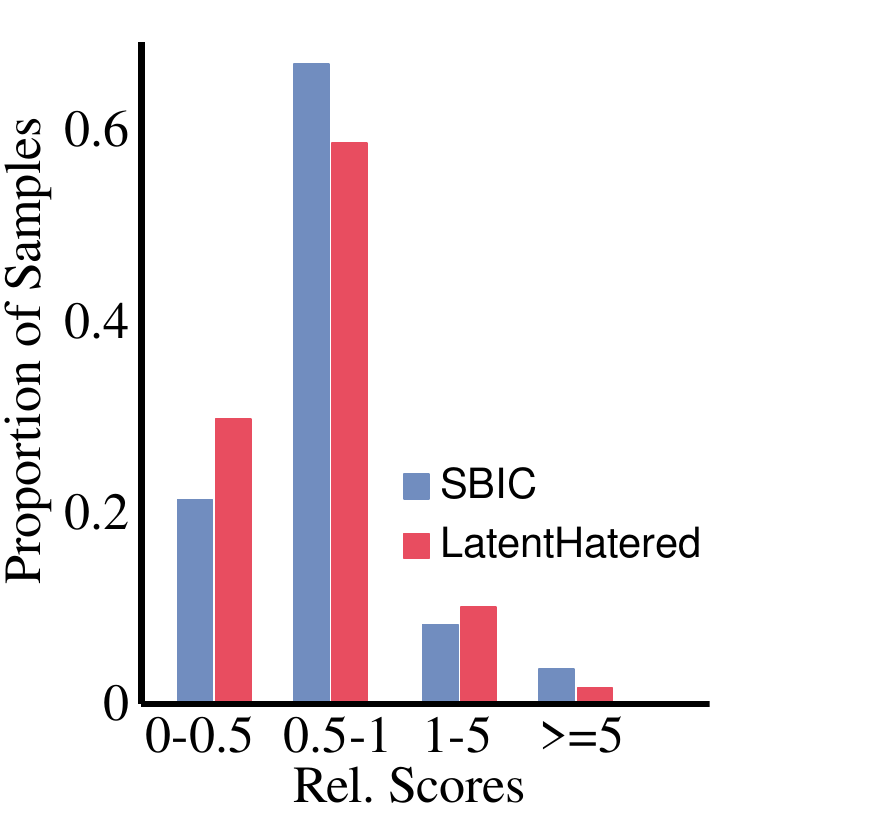}%
    }
    &
    \subfloat[\stereo]{%
        \label{d}%
        \includegraphics[width=0.28\textwidth]{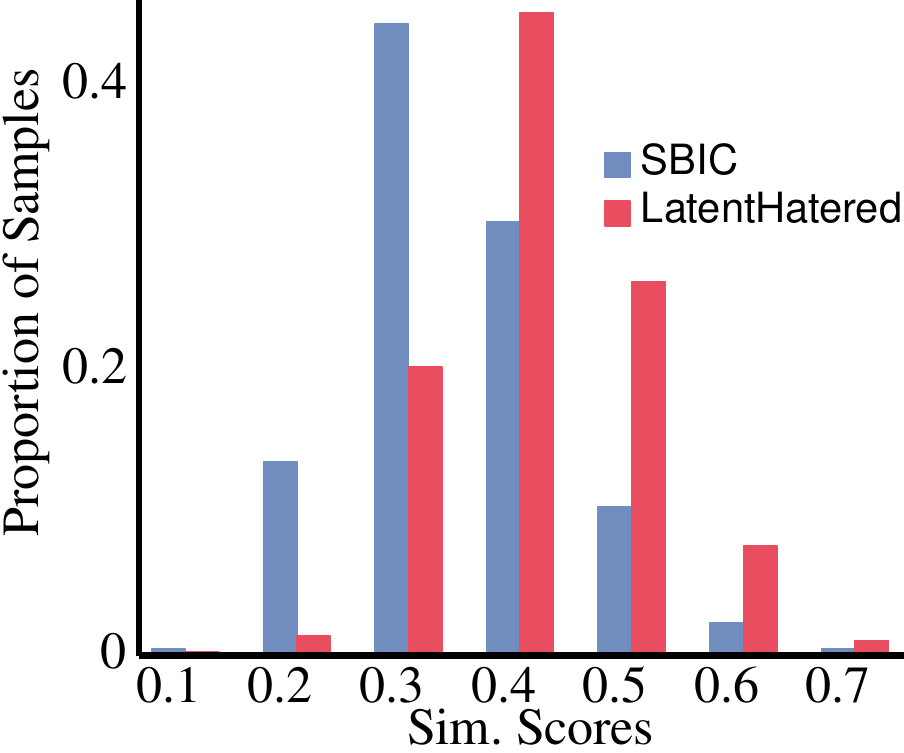}%
    }
\end{tabular}}
\caption{Analysis of top-k ($k=20$) KG tuples for \sbic\ and \latent\, capturing the spread of raw score values for (a) \concept\ and (b) \stereo\, respectively. Here, the x-axis represents the score value as either binned (for \concept) or rounded to the nearest 1st decimal (for \stereo). The bins range from [start, end) except for the last bin.}
\label{fig:kg_error}
\vspace{-3mm}
\end{figure}

\textbf{Manual assessment.} To ascertain the low quality of KG tuples, we manually inspect randomly selected $20$ samples from \sbic\ and \latent\ each. The corresponding top-k tuples are extracted w.r.t \concept\ and \stereo. Two expert annotators (details in Appendix \ref{app:kg_exp}) score each (input, top-k set) pair per KG. The manual labeling captures two components, `task-domain relevance' and `general-domain relevance,' scored separately on a $5$-point Likert scale. Task-domain relevance determines how effectively the retrieved tuples can explain implied stereotypes. A general-domain relevance determines if the tuples capture diverse concepts enlisted in the sentence from a common sense/world sense understanding. Table \ref{tab:manual_kg} lists the average (per annotator) scores and the inter-annotator cosine scores. We also observe a higher alignment of tuples in \latent\, which also explains the improvement in performance registered by this dataset under KG infusion (Table \ref{table:quality_cos_results}). As a toxicity-specific KG, \stereo\ seems to provide comparatively better tuples than \concept, yet both end up with abysmal relevance scores garnered by both annotators. \emph{Our manual inspection strongly corroborates that the quality of tuples is not informative/specific enough for our task.}

\textbf{Research Implications.} Despite their prominent use in NLP \cite{schneider-etal-2022-decade}, the question of analyzing the quality of KG-tuples needs to be explored at large. Our counter-intuitive observations and an examination of general purpose vs domain-specific KGs highlight the issue of signal/noise in the retrieved tuples. Our preliminary study paves the way for such analysis across NLP tasks. Our analysis shows that the defacto tuple retrieval filtering is not contextually sufficient to explain implicit hate. The absence of explicit hate or indirect mention of the target means that extracted entities may not relate to hateful connocations. 

Although language models positively exploit KG infusion \cite{chang-etal-2020-incorporating} to improve performance metrics, the KG infusions fall short of eliciting latent cognitive capabilities for social reasoning/subjective tasks such as implicit hate or sarcasm explanation. Similar issues in implicit hate detection tasks have been observed via automated evaluations \cite{lin-2022-leveraging}. However, ours is one of the initial work to look into this issue extensively. We suspect such behavior will occur in other NLP tasks as well. The work also calls for better infusion graph-based non-sequential information into seq-2-seq LLMs \cite{Besta_2024}. 

There is a need for domain-specific KG retrival and ranking methods of KG tuples. Regarding augmenting KGs, research in this area will benefit from efficient task-specific and multi-hop retrieval functions to enhance the quality of top-k tuples. Parallelly, there must be an active discussion on ``how LMs learn the association between external and pretrained features?"

\begin{table}[t]
\resizebox{\columnwidth}{!}{
\begin{tabular}{cc|cc|cc|cc}
\hline
\multirow{2}{*}{\textbf{D}} & \multirow{2}{*}{\textbf{KG}} & \multicolumn{2}{c|}{\textbf{A}\textsubscript{1}}       & \multicolumn{2}{|c|}{\textbf{A}\textsubscript{2}}       & \multicolumn{2}{|c}{\textbf{Cosine Sim.}} \\ \cline{3-8}
&  & \textbf{T}\textsubscript{r} & \multicolumn{1}{c|}{\textbf{G}\textsubscript{r}} & \textbf{T}\textsubscript{r} & \multicolumn{1}{c|}{\textbf{G}\textsubscript{r}} & \textbf{T}\textsubscript{r}  & \multicolumn{1}{c}{\textbf{G}\textsubscript{r}}  \\ \cline{1-8}

\multirow{2}{*}{\sbic} & C & \multicolumn{1}{r}{0.24 ($\pm$  0.44)} & 0.29 ($\pm$ 0.46) & \multicolumn{1}{r}{0.05 ($\pm$ 0.50)}  & 0.95 ($\pm$ 0.70)  & \multicolumn{1}{r}{0.47} & 0.56 \\
& S  & \multicolumn{1}{r}{0.43 ($\pm$ 51)}    & 0.43 ($\pm$ 51)   & \multicolumn{1}{r}{0.19  ($\pm$ 0.03)} & 0.52 ($\pm$ 0.36) & \multicolumn{1}{r}{0.52} & 0.68 \\ \hline
\multirow{2}{*}{\latent} & C & \multicolumn{1}{r}{0.3 ($\pm$  0.57)} & 0.65 ($\pm$ 0.75) & \multicolumn{1}{r}{0.2 ($\pm$ 0.41)}  & 0.4 ($\pm$ 0.60)  & \multicolumn{1}{r}{0.71} & 0.73 \\
& S  & \multicolumn{1}{r}{2.35 ($\pm$ 67)}    & 1.55 ($\pm$ 0.89)   & \multicolumn{1}{r}{1.15  ($\pm$ 0.59)} & 1.35 ($\pm$ 0.67) & \multicolumn{1}{r}{0.89} & 0.79 \\ \hline
\end{tabular}
}

\caption{Task (T\textsubscript{r}) and general domain (G\textsubscript{r}) relevance scores by annotators A\textsubscript{1} and A\textsubscript{2} on $20$ random \sbic\ and \latent\ samples. We report the mean (std.) scores. Cosine similarity captures the inter-annotator agreement w.r.t \concept\ (C) and \stereo\ (S).}
\label{tab:manual_kg}
\vspace{-3mm}
\end{table}

\section{Conclusion}
Having established the (ir)relevance of commonsense knowledge-based systems, we examine the efficacy of \textit{in-domain} and \textit{in-dataset} toxicity features. An in-depth evaluation also points out the expected behavior, which is that the random toxicity score does deteriorate the model's performance. Our error analysis highlights that subjective tasks mitigating toxicity cannot be fully automated. Here, the way forward is a human intervention to compile the final version of machine-generated labels and context. Future works must also focus on developing datasets and systems to enable social reasoning \cite{zhou-etal-2023-cobra} and reduce the inference cost of incorporating external signals by continued pretraining.
 
\newpage
\clearpage
\section{Acknowledgements}
Sarah Masud would like to acknowledge the support of the Prime Minister Doctoral Fellowship and
Google PhD Fellowship. The authors also acknowledge the support of our research partners, Wipro AI and IIIT-Delhi's Center for AI.

\section{Limitations}
From our study, it is evident that modeling implicit context is challenging for PLMs. Any toxicity analysis systems (whether classification or generation) suffer from social biases they learn from the extensive pretraining corpus and the subjectivity of the annotated downstream tasks \cite{10.1145/3580494}. This can induce implicit biases and be destructive in the long run \cite{gehman-etal-2020-realtoxicityprompts}. Incorrect identification of the target group or propagation of hallucinated stereotypes is equally problematic.
Further, given the implicit nature of the task, the proposed system may miss out on correctly identifying instances of sarcasm and irony. We also want to mention the cases of incomplete human annotations (not encompassing all viewpoints of the target group). The number of gold-label instances can be increased for each sample to accommodate more perspectives of the target community. Stereotyping and implicit hate datasets that capture contextual and cultural nuances beyond English (West) are largely missing. Lastly, it is essential to point out the dependency of the proposed model on the external toxicity signal (either manually annotated or obtained from an already finetuned endpoint).

\section{Ethical Considerations}
Our study uses publicly available datasets, open-source knowledge graphs, and PLMs, except GPT-3.5. Like any other hate speech-related artifact, our proposed system can be employed by nefarious elements to induce toxicity. Unmasking implicit hate by the nature of the task itself causes the generations to be explicit and potentially toxic. We argue that in the content moderation pipeline, this information is presented only to the content moderators and is not exposed to the users. The human subjects involved in the human evaluation of \toxbart\ and the inspection of KG-tuples are volunteer participants. No personal information of the subjects was saved during the evaluation phase.  

\bibliography{acl}

\newpage
\clearpage
\appendix
\section{Appendix}
\subsection{Experimental Details}
\label{appendix:exp_details}
\textbf{Engineering.} We use the HuggingFace Transformers Library \cite{huggingface-transformers} for our experiments, with BART Base \cite{lewis-etal-2020-bart} being our backbone network for the stereotype generation task, and BERT Base \cite{devlin-etal-2019-bert} being the model finetuned for the toxic attribute probability approximation task. To reiterate, the hidden state dimensions for the BART Base model are 768. For inference, following \cite{sridhar-yang-2022-explaining}, the length penalty hyperparameter was set to 5, and the number of beams for beam search was set to 10. The experiments are collectively performed over an NVIDIA RTX A5000 and A6000.  We also use \texttt{gpt-3.5-turbo} provided by OpenAI. 

\textbf{Data Preprocessing.} We follow the preprocessing pipeline adopted from \citet{sridhar-yang-2022-explaining}, where we replace NAN, URL, and special tokens. We lowercase the samples. The respective datasets already do initial masking of sensitive user information in \sbic\ and \latent. We do not perform any further masking.

\subsection{ToxicBERT}
\label{app:toxbert}
The RMSE scores on $D_{jigsaw}$ for the train and validation split are enlisted in Table \ref{tab:rmse}.

\begin{table}[ht]
\centering
    \begin{tabular}{c|c}
    \hline
    \textbf{Split}   & \textbf{Loss} \\\hline
        Train & 0.0592\\
        Validation & 0.06887\\ \hline
    \end{tabular}
    \caption{RMSE for the best checkpoint of \texttt{ToxicBERT}.}
    \label{tab:rmse}
\end{table}

\subsection{Additional Configurations for In-Domain Attributes}
\label{app:configs}
We discuss three additional configurations that have been studied for infusing the in-domain attributes.

\label{app:conf}
\textbf{Configuration 3 (C3).} We began with the very rudimentary concatenation of $\mathcal{P}$ with input $\mathcal{X}$. We first transform $\mathcal{P}$ into a higher dimension vector $\mathcal{\tilde{P}}$. This vector and the incoming posts are separately passed through the BART encoder, and the resultant latent embedding ($H_{toxic}$ and $H_{utter}$) are concatenated and passed through another linear transformation to downsize before feeding to the decoder. The set of Equations \ref{eq:c1} outlines the setup where $\mathcal{V}(.)$ refers to a linear transformation, and $[,]$ corresponds to the concatenation operation.
\begin{equation}
\begin{gathered}
    H_{toxic} = \mathcal{F_\theta}(\mathcal{V}_{6 \times d}(\mathcal{P})) ; \quad H_{utter} = \mathcal{F_\theta}(\mathcal{X}) \\
    \mathcal{\tilde{Y}} = \mathcal{G_\theta}(\mathcal{V}_{2\*d \times d}([H_{toxic}, H_{utter}]))
\end{gathered}
\label{eq:c1}
\end{equation}
Here, $H_{utter}$ and $H_{toxic}$ are the encoded representations of the input and the corresponding probability-to-special text tokens.

\textbf{Configuration 4 (C4).} We first to encode $\mathcal{\tilde{P}}$ and then concatenate. This will require the concatenated vector to undergo linear transformation to match the decoder dimension. The set of Equations \ref{eq:c3} outlines this setup.
\begin{equation}
  \begin{gathered}
    H_{toxic} = \mathcal{F_\theta}(\Gamma(\lambda, \mathcal{P})); \quad H_{utter} = \mathcal{F_\theta}(\mathcal{X}) \\
    \mathcal{\tilde{Y}} = \mathcal{G_\theta}(\mathcal{V}_{2\*d \times d}([H_{toxic}, H_{utter}]))
\end{gathered}
\label{eq:c3}
\end{equation}

\textbf{Configuration 5 (C5).} Building upon the previous configuration, here, instead of directly concatenating the two encoder outputs, we use the Compositional De-Attention framework (CoDA) \cite{quasi-attn}. CoDA determines the attention scores between the two encoder outputs. The intuition for this method is that some toxic attributes might be more critical or ``similar" for some token in the utterance than others, which can be considered ``dissimilar." The CoDA attention outputs are then combined with (via addition) encoder outputs of input utterances before passing through the decoder. Equations \ref{eq:c4} outline the setup.
\begin{equation}
    \begin{gathered}
    H_{toxic} = \mathcal{F_\theta}(\Gamma(\lambda, \mathcal{P})); \quad H_{utter} = \mathcal{F_\theta}(\mathcal{X}) \\
    \tilde{H} = H_{utter} + \psi(H_{toxic}, H_{utter}) ; \quad \mathcal{\tilde{Y}} = \mathcal{G_\theta}(\tilde{H})
\end{gathered}
\label{eq:c4}
\end{equation}

where $\psi$ refers to the CoDA framework \cite{quasi-attn} that captures the attention score via $ \psi = (tanh(\frac{Q \* K^\top}{\sqrt{d_{k}}}) \odot sigmoid(\frac{\Phi(Q, K)}{\sqrt{d_{k}}})) \* V$.

\subsubsection{Performance on Additional Configurations}
Table \ref{tab:ablation_exp_c1} shows that \toxbart\textsubscript{C3} performs worse than even vanilla BART and GPT-2. We conjecture this arises from the difference in the distribution space of probability scores vectors and BART representations. On the other end of the spectrum, we observe for  \toxbart\textsubscript{C5} that attentive concatenation may be overfitting the toxicity signals, leading to a loss of information. The lower efficacy of \toxbart\textsubscript{C5} aligns with previous research on attention-based KG-tuple concatenation \cite{sridhar-yang-2022-explaining}. Nevertheless, concatenation in the embedding space post encoding is not as effective as concatenation in the input space as in \toxbart\textsubscript{C1}.

\begin{table}[t]
    \centering
    \resizebox{\columnwidth}{!}{
    \begin{tabular}{lccc}
        \hline
        \textbf{Method} & \textbf{BLEU} & \textbf{ROUGE-L} & \textbf{BERTScore} \\\hline
        \toxbart\textsubscript{C1} & 64.89 & 63.83 & 64.52 \\ \hdashline
        \toxbart\textsubscript{C3} & 12.89 &  17.39 & 34.06\\
        \toxbart\textsubscript{C4} & 0.76 &  4.77 & 34.94\\
        \toxbart\textsubscript{C5} & 61.77 &  65.71 & 82.51\\\hdashline
    \end{tabular}}
    \caption{Ablations on \toxbart\textsubscript{C1} on \sbic for different in-domain configurations.}
    \label{tab:ablation_exp_c1}
\end{table}

\begin{table}[t]
\centering
\resizebox{\columnwidth}{!}{
\begin{tabular}{l|l}
\hline
\textbf{Token} & \textbf{Prompt}\\ \hline
$<TOXIC>$ & toxic \\
$<NOT\_TOXIC>$ & not toxic \\
$<SEVERE\_TOXIC>$ & severely toxic \\
$<NOT\_SEVERE\_TOXIC>$ & not severely toxic \\
$<OBSCENE>$ & obscene \\
$<NOT\_OBSCENE>$ & not obscene\\
$<IDENTITY\_ATTACK>$ & identity attack \\
$<NOT\_IDENTITY\_ATTACK>$ & no identity attack \\
$<INSULT>$ & insulting \\
$<NOT\_INSULT>$ & not insulting \\
$<THREAT>$ & threatful \\
$<NOT\_THREAT>$ & not threatful \\
\hline
\end{tabular}}
\caption{Token to Prompt mapping for ablation Exp 1.}
\label{tab:ablation_experiment_prompts}
\end{table}

\subsection{Human Evaluation of \toxbart}
\label{app:human_eval}
Here, we provide details about the process of engaging the human evaluators, the annotation guidelines, and a note on target identification. 

\subsubsection{Evaluator Recruitment} 
As stated in Section \ref{sec:exp_rq2}, we recruit $20$ human evaluators aged $18+$ who have experience in using social media and work in computational social science and natural language processing. The evaluation is voluntary, with no monetary compensation. It should be noted that while the initial shortlisting of samples to annotate was random, the final samples for evaluation were selected by the authors after vetting the initial text and its ground explanation (without looking at any model output) to minimize risk and harmful exposure for human subjects. We attempted to be as fair and diverse in our selection of samples as possible. Before the evaluation, we reached out to the people interested in participating. We gave a detailed overview of the task (via email), providing them with material to sensitize them towards the task at hand.
Further, the reviewers were known to participate in some hate speech-related evaluations prior and had an idea about the content they would be engaging with. Only those willing to participate participate consensually were invited for the review. Apart from the warning posted in the Google form, the evaluators were encouraged to contact the authors anytime during their evaluation to share feedback or discuss the content.

\subsubsection{Annotation Guidelines}
The evaluators are provided the following information blob and are free to reference the information anytime during their assessment. With a range of 1-5, the user is not forced to select/rank between the two. They can access the results independently for both systems. 

\begin{tikzpicture}
\draw [thick,dash dot] (0,1) -- (7,1);
\end{tikzpicture}

Kindly go through the points below to gain context about the task before filling out the Google form. Filling the form out should not take more than 20-25 minutes. Thank you for your time!

\textbf{Note:} This form contains content that some might find offensive and upsetting. Reader discretion is advised.

\textbf{Terminology:}

\textbf{Stereotype:} According to the Wikipedia article, a stereotype is referred to as "a generalized belief about a particular category of people." 

\textbf{Stereotypical utterance:} A stereotypical remark is an utterance that indirectly/implicitly hints at a stereotype.

\textbf{Implied Stereotype:} A short explanation in free text form of the stereotypical remark expressing the negative and often offensive intent behind the remark towards the target group/category of people.

For each utterance (which may or may not be hateful), there are two machine generations for the implied stereotype expressing the intent behind the utterance. Each utterance will be referred to by the code U{x}, where {x} is some number from {1-10} and the first generation by S{x}a whereas the second generation by S{x}b. For example, U3 refers to Utterance \#3, S3a refers to the first stereotypical implication generation, and S3b refers to the second generation.

For each generation, there are five metrics you will have to evaluate. We follow the 5-point Likert scale, with five being the highest. One metric is on a binary scale. You are required to compare each generation with the corresponding utterance and answer the questions which follow accordingly.

\begin{enumerate}[noitemsep, nolistsep, topsep=0pt, leftmargin=1em]
\item \textbf{Fluency} measures how fluent the generation is in English, irrespective of its context regarding the task and its corresponding utterance. We only consider the syntactic properties of language here. Example: ``My name is John'' is a fluent sentence.

\item \textbf{Coherency} measures how coherent the generation is. This is with respect to the utterance and the task. We aim to look at only the syntactic features via this metric. Example: Given an utterance that makes a stereotype against black folks indulging in criminal activities, the generation ``this is a racial stereotype'' is coherent with the utterance because it grabs the correct context regarding the utterance. Whereas generation like ``mentally disabled folks are dum'' is not because the original utterance is not talking about mentally disabled folks.

\item \textbf{Specificity} measures how specific the generation is when considering the context of the utterance. This metric also determines how much contextually specific information is present in the generation, but not the correctness. We aim to look at the semantic correctness via this metric. Example: For the same utterance as for the previous metric, the generation ``this is a racial stereotype against black folks indulging in criminal activities'' is much more specific than ``this is a racial stereotype.'' Both generations might be equally coherent, but that does not imply how specific they are.

\item \textbf{Similarity with gold explanation} Similarity with gold explanation determines how similar the generations are with respect to any of the given gold annotations. You can combine your observations from metrics 2 and 3 here. Example: Given the gold label ``racial stereotype against black folks indulging in criminal activities.'' The generation ``this is a racial stereotype against black folks'' is much more similar to the gold label than ``this is a racial stereotype''.

\item \textbf{Target Group} determines how correctly the generations identify the target group. You will be provided with the gold label and asked to mark whether the stereotype targets the same group. Option 0 [Target Not Correct] will be the valid option if the generation does not seem to target any group.
\end{enumerate}
\begin{tikzpicture}
\draw [thick,dash dot] (0,1) -- (7,1);
\end{tikzpicture}

\subsubsection{Note on Target Group} To clarify, we did not explicitly prompt any model under examination to separately predict the target. Instead, human evaluators determine if the model under evaluation can detect the correct target group within the explanation it generates. Here, we observe that human evaluators found that 48\% of the time, GPT-3.5 focused on either the wrong target group or talking about the wrong stereotype for the given target group. We want to point out that the target group specified in both datasets is annotated by humans in the respective datasets in a free text form, leading to some raw 800 different target names. A categorical detection and assessment are not possible feasible. Hence our reliance on human evaluation.

\begin{table}[!t]
\resizebox{\columnwidth}{!}{
\begin{tabular}{l|p{8em}|p{8em}}
\hline
\textbf{KG/Property}      & \textbf{ConceptNet}                    & \textbf{StereoKG}\\\hline
Size (\# tuples)          & $\sim$34M                              & $\sim$4k                       \\
Curated from       & Wikipedia                              & Reddit (offensive subreddits)     \\
Type of tuples            & World and commonsense knowledge & Religious and ethnic stereotypes \\
Top-k tuples via & Weighted TF-IDF                       & Cosine Similarity \\\hline
\end{tabular}
}
\caption{Summary comparison of the properties of the KGs involved in our investigation}
\label{tab:kg_prop}
\end{table}

\begin{figure}[!t]
    \subfloat[\concept]{%
        \label{a1}%
        \includegraphics[height=0.24\textwidth]{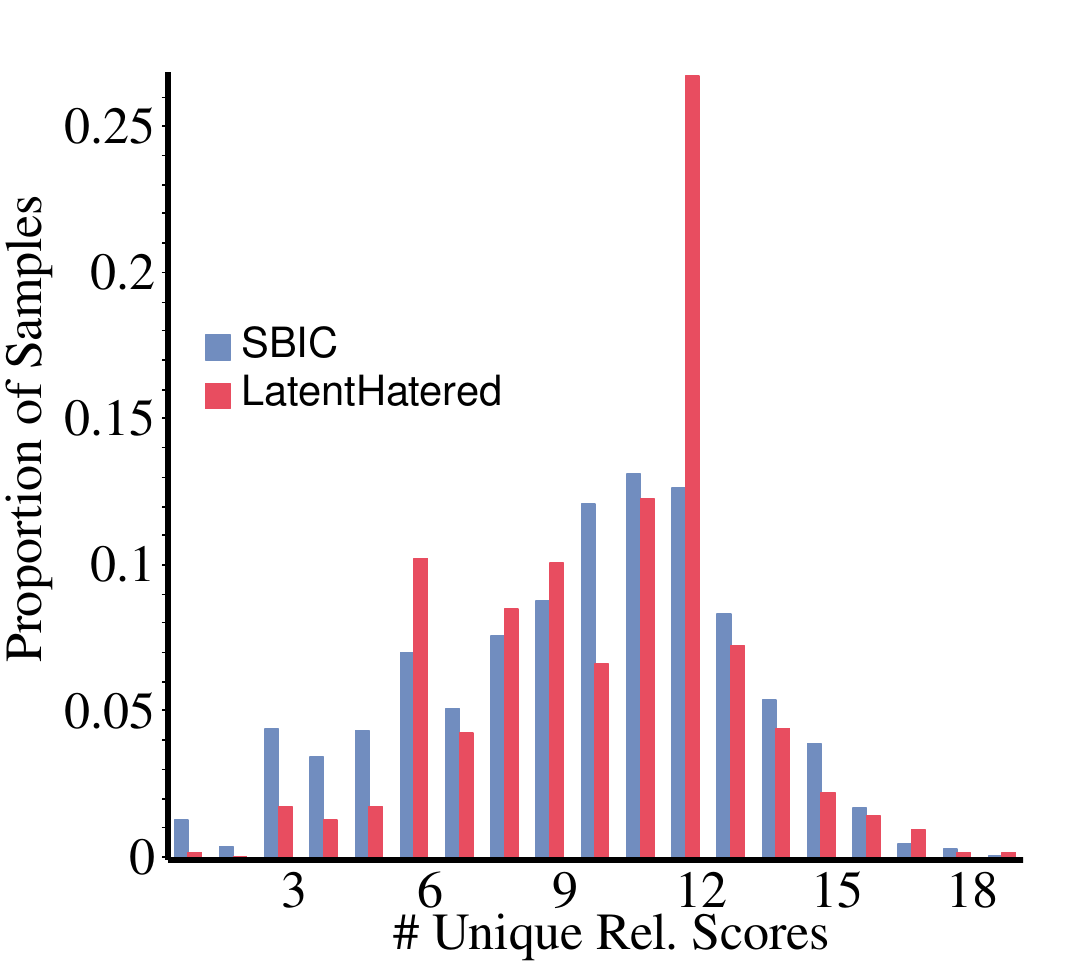}%
    }
    \subfloat[\stereo]{%
        \label{b}%
        \includegraphics[height=0.22\textwidth]{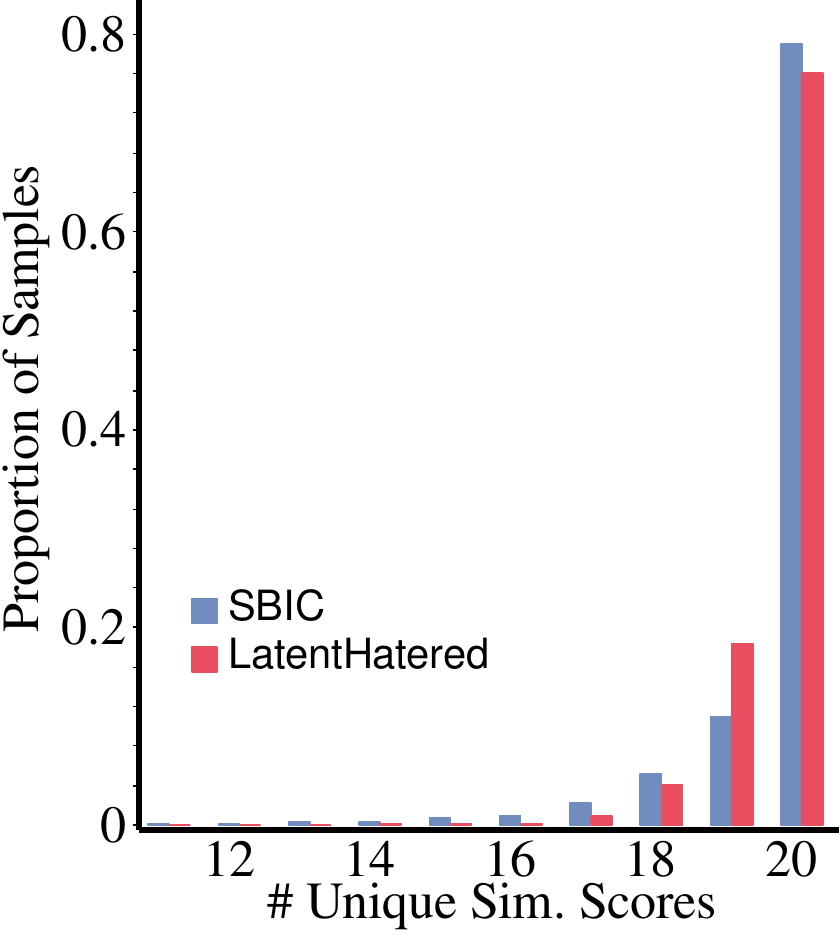}%
    }
\caption{Analysis of top-k KG tuples retrieved for test samples of \sbic\ and \latent\ at $k=20$, w.r.t \concept, and \stereo. as described in Section \ref{section:cos_quality_relevance}, for \concept\, we evaluate the IDF weighted relevance (rel.) scores. For \stereo, we evaluate via the cosine similarity (Sim.) scores. All the y-axis captures the proportion of samples corresponding to the analysis at hand. Given that we look at top $20$ tuples based on scores, (a) and (b) capture the spread of uniqueness in scores obtained per sample, respectively, for \concept\ and \stereo\. Here, the $i$th index on the x-axis is the number of unique scores out of $20$ present in the samples.}
\label{fig:kg_error_app}
\end{figure}

\subsection{Auditing KG Attributes}
\label{app:kg_exp}
\textbf{Choosing k-relevant tuples.} For \concept, we follow the retrieval method used by \citet{chang-etal-2020-incorporating} and \citet{sridhar-yang-2022-explaining}. In this setup (\textit{Algorithm 1}), we first obtain the query terms ($q$) from the input post's lemmatized noun, verb, and adjective keywords. We then extract from \concept\ all the 1-hop English tuples for each term. We also calculate the IDF score for each query term, $idf_{q}$. The top-k and bottom-k tuples are obtained by sorting the extracted relations based on relevance scores $W_{rel} \times idf_{q}$, as each relation in \concept\ has a \emph{relation-weight}, $W_{rel}$. For random-k, we randomly pick $k$ tuples from the extracted set.

For \stereo, we utilize semantic similarity-based metric (\textit{Algorithm 2}). We first employ the \texttt{all-MiniLM-L6-v2} \cite{reimers-2019-sentence-bert} to pre-calculate the sentence embeddings over all the linearised\footnote{The linearised tuples are already provided along with the triplets at: https://github.com/uds-lsv/StereoKG/} tuple from \stereo. We then used the cosine-similarity scores between tuples and input samples to get the top and bottom-k tuples. The same algorithm cannot be applied to both KGs due to the skewness in KG size. 

Employing \textit{Algorithm 1} for \stereo\ returns a low (and zero in most instances) number of tuples per input sample. Meanwhile, employing \textit{Algorithm 2} for querying on \concept\ is not computationally feasible as it amounts to performing cosine similarity in the order of millions. Further, following the experimental setup from MIXGEN \citet{sridhar-yang-2022-explaining}, we also set $k=20$. The pseudo codes for the tuple extraction via the respective KG are outlined in Algorithm \ref{algo:1} and \ref{algo:2} for \concept\ and \stereo\, respectively. 

\begin{algorithm}
\caption{Knowledge Tuples Extraction for ConceptNet}\label{alg:conceptnet}
\label{algo:1}
\begin{algorithmic}[1]
\Ensure $KG_h = \{(r_i, t_i, score_i) \mid 0 \leq i \leq N_i \}$
\State $query\_tokens \gets$ extract adjectives, nouns, and verbs from each post
\State $idf\_scores \gets$ TF-IDF scores of each query token given the vocabulary of all posts \\
rank relevant tuples from $KG_h$ in terms of $idf\_scores_h \cdot score_i$, where h is a query token
\end{algorithmic}
\end{algorithm}

\begin{algorithm}
\caption{Knowledge Tuples Extraction for StereoKG}\label{alg:stereokg}
\label{algo:2}
\begin{algorithmic}[1]
\Ensure $KG_h = \{(r_i, t_i, score_i) \mid 0 \leq i \leq N_i \}$  
\State $emb\_vec \gets$ embedding of each post from model $\mathcal{Q}$
\State $lin\_KG \gets$ Linearised tuples from \stereo
\State $cosine\_sim(emb\_vec,emb\_vec)$
\State rank relevant tuples in terms of $cosine\_sim$
\end{algorithmic}
\end{algorithm}

\textbf{Relevancy Scoring.} In Figure \ref{fig:kg_error_app}, we look at the number of unique scores (relevance or similarity for \concept\ and \stereo\ respectively) obtained for a sample. For $k=20$, one would expect the uniqueness to be right-skewed, which is partially valid for \stereo\ but not for \concept\ where there are fewer samples with $>=16$ unique scores and zero samples with all unique scores. Interestingly, despite the similarity metric being limited to $0-1$ for \stereo\, it produces a higher number of unique scores compared to \concept. The relevance metric is open-ended $\geq0$ for the latter. One would expect that an open-ended metric will generate more variation in scores. However, this is not the case.

\textbf{Annotator Demographic for Manual Inspection.} To manually examine the KG tuples, we took help from $2$ expert annotators who volunteered $\approx35$ minutes each and scored $20$ samples and their $top-k=20$ KG tuples. The annotators, one male (24 years) and one female (29 years), are knowledgeable about natural language processing and social computing. Additionally, both adequately understand how KG's are constructed and employed in NLP. Besides providing scores, the annotators could offer any additional comment about an outlier they observed. 
\end{document}